%% file: output.tex
\pdfoutput=1
\documentclass[11pt]{article}

\usepackage[]{acl}

\usepackage{times}
\usepackage{latexsym}
\usepackage[T1]{fontenc}
\usepackage[utf8]{inputenc} 
\usepackage{microtype}
\usepackage[defaultcolor=magenta]{changes}
\usepackage{multirow, multicol} 
\usepackage{makecell} 
\usepackage{adjustbox} 
\usepackage{amsmath} 
\usepackage{pifont} 
\newcommand{\cmark}{\ding{51}} 
\newcommand{\xmark}{\ding{55}} 
\usepackage{enumitem} 
\newcolumntype{C}[1]{>{\centering\arraybackslash}p{#1}} 
\usepackage{hyperref} 
\usepackage{ulem} 

\title{EHR-SeqSQL : A Sequential Text-to-SQL Dataset \\For Interactively Exploring Electronic Health Records}

 \author{Jaehee Ryu$^*$,  Seonhee Cho$^*$,  Gyubok Lee,  Edward Choi \\
        KAIST \\ \texttt{\{jh.ryu, seonhee.cho, gyubok.lee, edwardchoi\}@kaist.ac.kr} }

\begin{document}
\maketitle

\def\thefootnote{*}\footnotetext{These authors contributed equally to this work.}\def\thefootnote{\arabic{footnote}}

\begin{abstract}
In this paper, we introduce EHR-SeqSQL, a novel sequential text-to-SQL dataset for Electronic Health Record (EHR) databases. EHR-SeqSQL is designed to address critical yet underexplored aspects in text-to-SQL parsing: interactivity, compositionality, and efficiency. To the best of our knowledge, EHR-SeqSQL is not only the largest but also the first medical text-to-SQL dataset benchmark to include sequential and contextual questions. We provide a data split and the new test set designed to assess compositional generalization ability. Our experiments demonstrate the superiority of a multi-turn approach over a single-turn approach in learning compositionality. Additionally, our dataset integrates specially crafted tokens into SQL queries to improve execution efficiency. With EHR-SeqSQL, we aim to bridge the gap between practical needs and academic research in the text-to-SQL domain. EHR-SeqSQL is available at \url{https://github.com/seonhee99/EHR-SeqSQL}.


\end{abstract}

\begin{figure}[t!]
\centering
    \includegraphics[width=7.4cm,height=13.2cm]{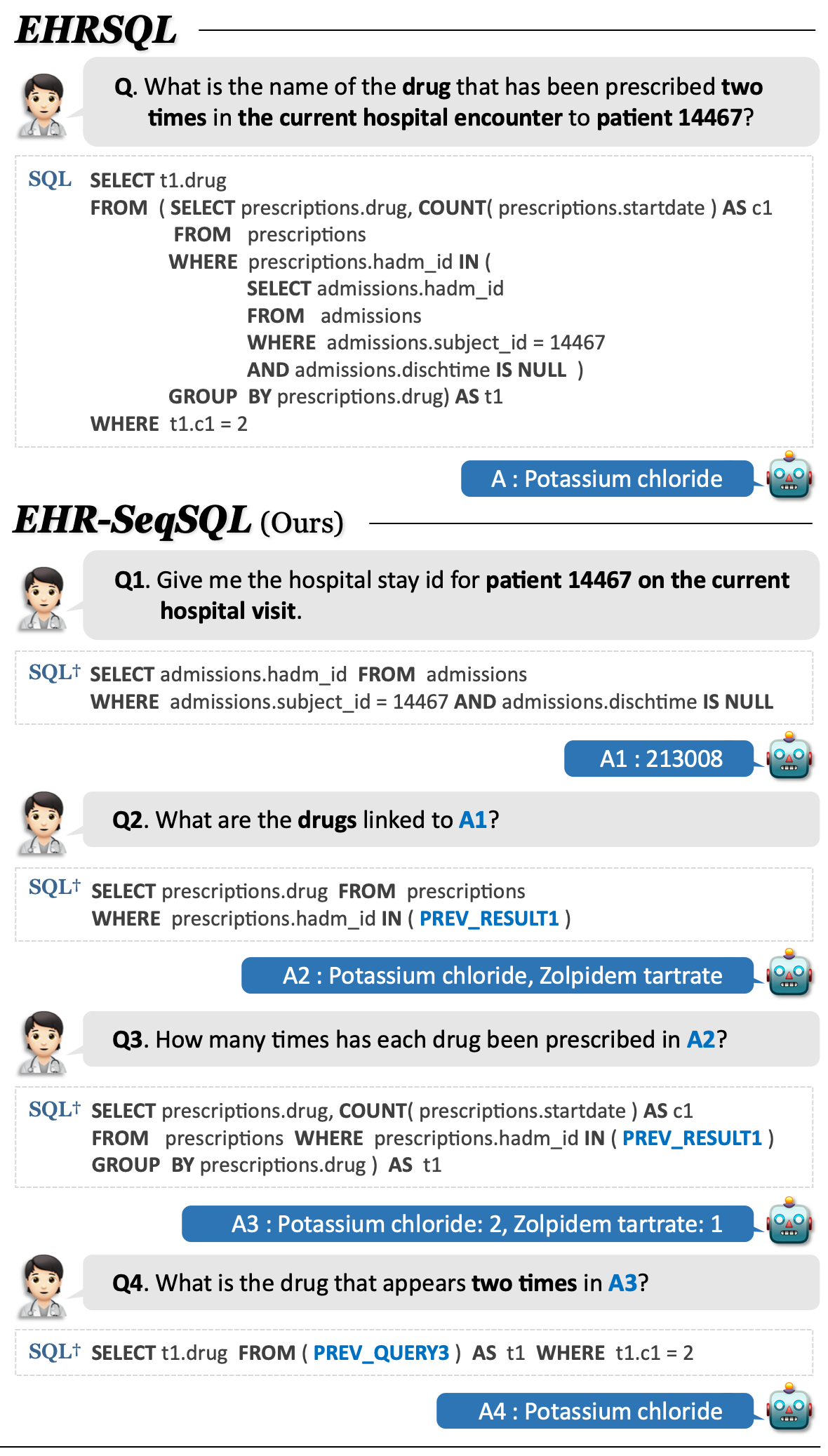}
    \caption{
    \textbf{EHRSQL vs. EHR-SeqSQL }
    EHR-SeqSQL is a dataset that adapts the single-turn setting of EHRSQL into a multi-turn setting. The SQL queries in EHR-SeqSQL include the special tokens to refer to the previous context, which can be executed in the database with simple post-processing.}
    \label{figure1}
\end{figure}

\section{Introduction}

Text-to-SQL provides a practical opportunity for non-experts to explore databases, even without prior knowledge of the database operations.
Electronic Health Records (EHRs) are large-scale relational databases (RDBs) storing vast and comprehensive patient data~\cite{johnson2016mimic, pollard2018eicu}.
Medical experts often ask questions that require highly complex reasoning across multiple tables and access to a vast number of records within a single query~\cite{ehrsql}.
Handling such complexity in large-scale databases remains a significant challenge in the current text-to-SQL research~\cite{BIRD}.

Several efforts have been made to construct text-to-SQL datasets for EHRs. MIMIC-SQL~\cite{wang2020text} is the first text-to-SQL dataset that targets a subset of MIMIC-III~\cite{johnson2016mimic}, one of the widely-used open-source EHR databases, consisting of 26 tables.
DrugEHRQA~\cite{bardhan2022drugehrqa} provides the first question answering dataset that incorporates both structured tables and unstructured notes from EHRs.
EHRSQL~\cite{ehrsql} is a dataset curated based on a survey from various medical experts, reflecting the diverse information needs of the actual medical field.

Still, there are important yet underexplored objectives for the practical application of text-to-SQL models.
\textbf{First, the text-to-SQL system should incorporate interactivity.}
Most of the existing text-to-SQL research assumes a single-turn scenario, whereas the process of exploring information in real-world situations is often continuous~\cite{SQA, COSQL, SPARC}. For better user usability, it is desirable for the system to be interactive as well. 
\textbf{Second, the text-to-SQL system should learn compositionality.}
Realistically, datasets cannot capture all the diverse needs of users. Therefore, it is crucial that the model can handle a wider range of queries especially when these unseen queries comprise components (\textit{i.e.} sub-queries) that the model has explicitly seen during training.
This is the compositional generalization ability, which is challenging even for the highly proficient language models \cite{qiu2022evaluating}.
\textbf{Lastly, the text-to-SQL system should consider efficiency.}
Real-world databases are often significantly larger than academic ones \cite{hazoom2021sede, zhou2021learned}. Efficient query execution is crucial, given the magnitude of the databases in actual hospitals (\textit{e.g.} MIMIC-III dataset includes the medical history of 46k patients). This becomes even more significant when we consider a real-time interaction scenario between a text-to-SQL model and a user \cite{BIRD}. 

In this work, we introduce EHR-SeqSQL which addresses all three objectives. Our contribution is as follows:
\begin{itemize}[leftmargin=5mm]
    \item \textbf{Construction of a sequential text-to-SQL dataset for exploring EHR data:} As illustrated in Figure \ref{figure1}, EHR-SeqSQL is designed for multi-step interactions, built by decomposing the diverse and complex queries of EHRSQL and setting each question as the interaction goal.
    To the best of our knowledge, EHR-SeqSQL is the first multi-turn text-to-SQL dataset targeting structured medical records.
    We make our dataset public to encourage future research.

    \item \textbf{Validating the effectiveness of EHR-SeqSQL in the compositional generalization:} We designed two experiments to evaluate the compositional generalization ability.
    The results show that decomposing questions enables the model to better generalize to unseen interaction goals during training. 
    Furthermore, it demonstrated the potential for generalization on longer sequences not encountered during training.
    
    \item \textbf{Proposing special tokens for SQL:} We introduce a highly effective way to enhance query execution efficiency in multi-step interactions with the use of the novel special tokens for SQL.
    These tokens significantly increase time efficiency during query execution as well as improve the performance of text-to-SQL models. 
    These tokens are not only domain-agnostic but also more effective as the database size increases.
\end{itemize}

\begin{figure*}[t!]
    \centering
    \includegraphics[width=15.9cm,height=10.1cm]{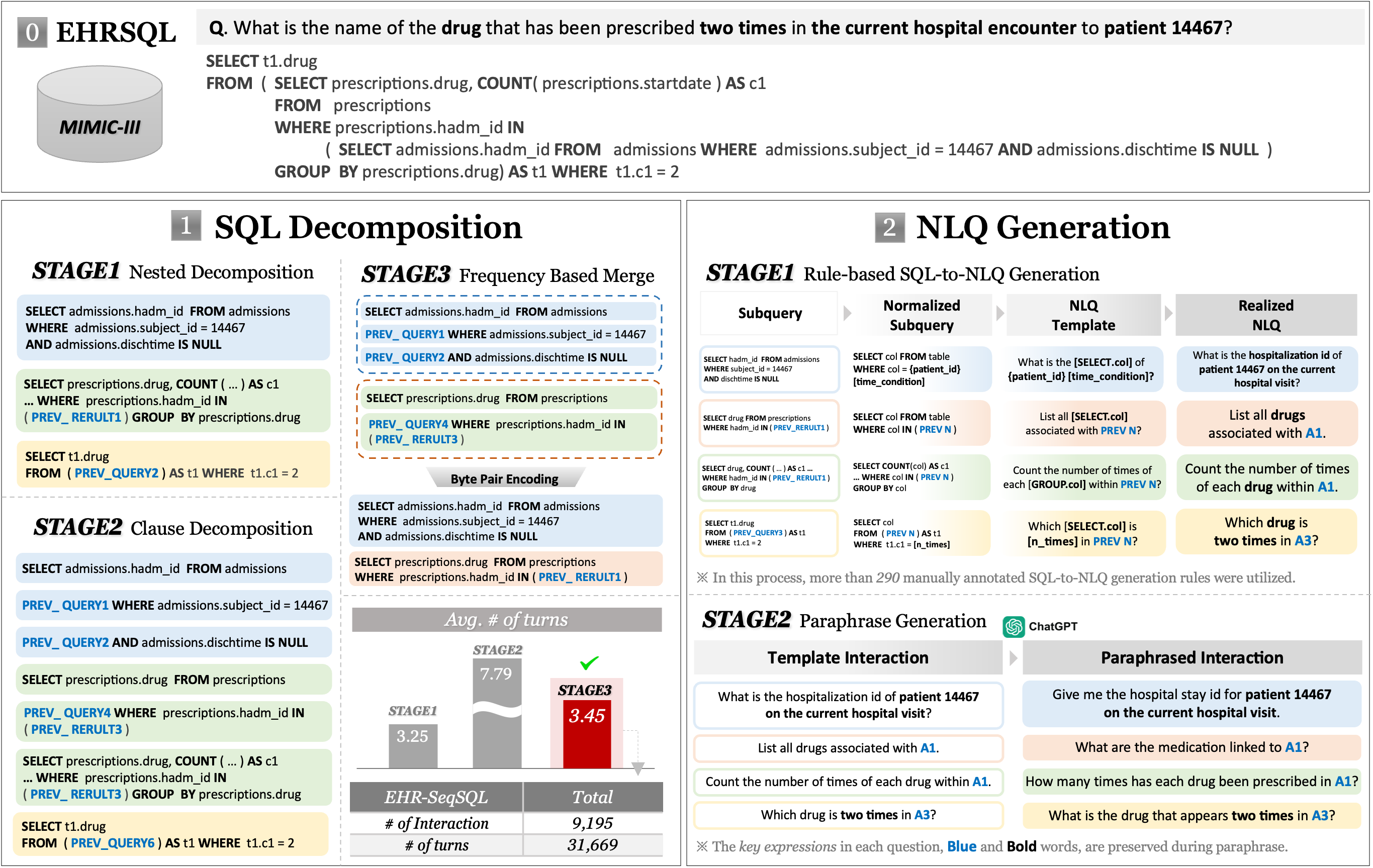}
    \caption{\label{figure2}
    Overview of the dataset construction process.
    We transform EHRSQL's single text-SQL pairs into multi-turn pairs for EHR-SeqSQL by first breaking down the original SQL into subqueries (Stages 1 and 2), then merging common patterns with the BPE algorithm (Stage 3). Natural language questions (NLQs) are created for each subquery using templates and paraphrased for clarity using ChatGPT.
    }
\end{figure*}

\section{Related Work}
\subsection{Multi-turn Text-to-SQL}

There are only a few datasets that are specifically designed for context-dependent text-to-SQL tasks. ATIS~\cite{atis} was the first to include a series of user questions aimed at interacting with a flight database. 
Recently, \citet{COSQL, SPARC} proposed cross-domain context-dependent text-to-SQL datasets, based on the questions from Spider~\cite{yu2018spider} serving as interaction goals.
While their motivation aligns closely with ours, their questions and the grounding databases are predominantly simplistic. In real-world scenarios, however, users typically have much more complex requirements, and databases often contain a significantly larger number of rows~\cite{ehrsql, BIRD}.
EHR-SeqSQL is designed upon this objective, based on the questions collected from an actual survey and grounding the real-world database, MIMIC-III~\cite{johnson2016mimic}. To the best of our knowledge, this is the first multi-turn text-to-SQL dataset in the healthcare domain.

\subsection{Compositional Generalization}
Compositional generalization, a major challenge in semantic parsing, enables a model to manage diverse and unfamiliar user queries using components recognized from training. It is typically achieved by creating different training and test splits of interest. \citet{SCAN} discovered that sequence-to-sequence models struggle to learn compositional structures, as demonstrated through their primitive and length splits. More recently, template splits~\cite{finegan2018improving} and TMCD splits~\cite{tmcd} were proposed to evaluate compositional generalization on text-to-SQL datasets. Notably, recent efforts~\cite{gan-etal-2022-measuring,liu2023exploring} have proposed compositional generalization in multi-turn settings, which aligns with the concepts we use in this paper. However, we aim to extend their approach by proposing and experimenting with two different challenges in multi-turn compositional generalization: unseen (Section \ref{sec:exp1}) and longer (Section \ref{sec:exp2}) interactions.

\section{Data Construction}
\label{sec:data_construction}
We aim to convert text-SQL pairs of EHRSQL to multi-turn text-SQL pairs that embrace interactivity and query execution efficiency.
Note that EHRSQL contains text-SQL samples for two different EHR sources, namely MIMIC-III~\cite{johnson2016mimic} and eICU~\cite{pollard2018eicu}.
We use the MIMIC-III version of EHRSQL, given its complex schema and wider adoption by the NLP community.
Questions for eICU and the questions that cannot be transformed into a proper SQL query are not the scope of this dataset but can be addressed in the future work.

Our data construction process is divided into two stages: SQL decomposition and natural language question (NLQ) generation. Initially, we break down each SQL query of EHRSQL into a sequence of subqueries to create more granular meanings of queries. Next, we create a corresponding NLQ for each subquery. Compared to NLQ decomposition, SQL decomposition enables systematic breakdown of a question without any loss of information, due to the strict, rule-based nature of SQL syntax.
The overall process is shown in Figure \ref{figure2} and we provide detailed explanations in the following sections.

\begin{table*}[h]
\centering
\begin{tabular}{c|c|c|c|c}
\hline
\multicolumn{1}{l|}{} & ATIS & SParC & CoSQL                & \textbf{EHR-SeqSQL}           \\ \hline
\# of Interactions    & 1,658         & 4,298          & 3,007 & 9,195 \\ 
\# of Turns           & 11,653        & 12,726         & 15,598                        & 31,669                        \\ 
Avg. \# of Turns       & 7.0             & 3.0              & 5.2                           & 3.5                          \\ 
Avg. \# of Tables / DB             & 27            & 5.1          & 5.1 & 26                            \\ 
Domain                & Airline       & Cross          & Cross                         & Medical                        \\ 
Compositional Split                & \xmark       & \xmark          & \xmark                         & \cmark                       \\ 
SQL Execution Efficiency                 & \xmark       & \xmark          & \xmark                         & \cmark                        \\ \hline
\end{tabular}
\caption{Comparison of EHR-SeqSQL with other multi-turn text-to-SQL datasets.}
\label{tab:data_statistics}
\end{table*}

\subsection{SQL Decomposition}
\label{sec:3.1}

\paragraph{Stage 1. Decomposing Nested Query}
\label{sec:3.1.1}

Since most SQL queries in EHRSQL have a nested structure, we decompose the queries based on their nesting levels. This method allows each subquery to contain more granular meanings of the original query and facilitate the multiple turns of interaction. We start by asking the inner query first, and then the outer query is asked while referring to the result of the inner query. This decomposition approach guarantees the executability of queries at each turn and allows anaphoric expressions in the subsequent turns.

In addition, we devised two special tokens for SQL to leverage the previous turn's subqueries and results: \texttt{prev\_query} and \texttt{prev\_result}. These tokens are accompanied by a specific turn index (e.g. \texttt{prev\_query1}, \texttt{prev\_result2}). 
Specifically, \texttt{prev\_query} token refers to the generated query of the specified turn, whereas the \texttt{prev\_result} token refers to the execution result of the specified turn. 
Two tokens are substituted with either the referred query or its execution result before execution. The effectiveness of these tokens is later discussed in Section~\ref{sec:exp3}.

\paragraph{Stage 2. Decomposing SQL Clauses}
\label{sec:3.1.2}
After Stage 1, multi-turn questions are obtained, where each can be asked with reference to the answer to previous questions.
Still, these questions often convey multiple conditions that users might ask in separate questions. For example, the question ``\textit{What was the last hospitalized admission time that patient 17694 was admitted via transfer from emergency room?}'' contains the specific conditions about the admission time, the patient, and the admission route.
Therefore, we further decompose queries on the clause-level: WHERE, ORDER BY, and HAVING clauses and aggregation functions (\textit{e.g.} \verb|MAX, MIN, SUM|) in the SELECT clause. 
In cases where multiple clauses appear together, we parse in the order of SQL execution.
Exceptions were made only when the original semantics of the query became ambiguous after being decomposed.
More details are in Appendix \ref{appendix:stage2_exception}.

\paragraph{Stage 3. Merging Subqueries by Frequency}
\label{sec:3.1.3}
We noted that frequently appearing consecutive subqueries are most likely decomposed due to the specific database schema characteristics or SQL structures rather than each being queries users would naturally ask.
In Stage 3, we employ a recursive application of the Byte Pair Encoding (BPE) algorithm to merge frequent consecutive subquery pairs. This recursive approach allows for the amalgamation of not only two but up to three or more subqueries, thereby simplifying complex patterns and enhancing query interpretability. Further details are given in Appendix \ref{appendix:stage3_details}.

\subsection{NLQ Generation}
\label{sec:3.2}

\paragraph{Stage 1. Rule-based SQL-to-NLQ Generation}
While previous SQL-to-NLQ studies~\cite{spiderss, wu2021data} decomposed SQL at the clause level for sub-questions and concatenated clauses later, such simple concatenation has a risk of unnaturalness.
Thus we create NLQ templates for each corresponding SQL subqueries for the quality of NLQ.
To efficiently annotate NLQs for each corresponding SQL subqueries, we first normalize subqueries by replacing specific table names, column names, and condition values to abstract terms such as \texttt{table} and \texttt{col} (see Figure~\ref{figure2}).
We then manually create NLQ templates for each normalized subqueries.
These templates include slots for table names, column names, condition values, SQL functions (\textit{e.g.} GROUP BY, AVG), and time expressions.
Then the actual question is generated from the NLQ templates by the slot-filling process. 
More details are in Appendix \ref{app:NLQ}.

\paragraph{Stage 2. Paraphrase Generation}
We further paraphrase the NLQ templates to enhance linguistic variability within the dataset, using ChatGPT for its superior performance in understanding and generating text \cite{guo2023close}.
We ensured that all the slot values and turn indices were preserved after paraphrasing.
Additionally, we employed the self-consistency method and a duplicate question detection model to ensure the quality of the paraphrases, following \citet{ehrsql}.
On average, we obtained 10.39 paraphrases for each NLQ template. 
Finally, we randomly assign the paraphrased template to each question and fill the slots with the original condition values.

\paragraph{Quality Check}
The authors meticulously conducted two-step quality check at both the turn level and the interaction level. All question templates used during the generation process were checked against the following criteria: \textbf{1) Completeness}: All information in the SQL is also explicitly stated in the NLQ, and \textbf{2) Naturalness}: Questions are formulated to be as natural as possible after the template masks are realized. Every question and SQL template was crafted to meet these standards. We then conducted quality checks on 1,000 randomly selected interactions from the final dataset at both the turn and interaction level. At the turn level, we checked for clarity of each question on the turn level, since they contain coreferences. At the interaction-level, whether the interactions accurately represent the original intent of the EHRSQL questions is evaluated. We revised the templates or introduced more templates wherever issues were identified. This process was repeated until every sampled interaction met our stringent quality standards at both the turn and interaction levels.

\subsection{Data Statistics}
Table~\ref{tab:data_statistics} presents the statistics of multi-turn text-to-SQL datasets.
EHR-SeqSQL has the largest number of interactions and turns compared to all existing multi-turn text-to-SQL datasets. 
Notably, this is the first work to introduce special tokens within SQL queries to improve execution efficiency. Furthermore, we also provide a new split and additional test set to evaluate the compositional generalization, as detailed in Section \ref{experiment}.

Questions in our dataset are categorized into four types: \textit{independent}, \textit{referential}, \textit{filtering}, and \textit{modifying}. These categories are not mutually exclusive and a question may belong to more than one category. \textit{Independent} questions are those without any prior context. \textit{Referential} questions refer to previous questions or answers while \textit{filtering} questions narrow the previous question's scope by adding conditions. \textit{Modifying} questions are particularly challenging as they only mention the altered condition and omit all the same conditions. Table \ref{tab:data_category} shows the distribution of the questions.
\begin{table*}[t!]
\centering
\begin{tabular}{c|ccc|ccc}
\hline
\multirow{2}{*}{\textbf{Split}} & \multicolumn{3}{c|}{\textbf{Random}} & \multicolumn{3}{c}{\textbf{Compositional}} \\ \cline{2-7} 
& Train & Test & Test$_{L}$  & Train  & Test & Test$_{L}$    \\ \hline
\renewcommand{\arraystretch}{1}
\# of Question Temp   & 167     & 166  & 166        &  121    & 46     & 46      \\ 
\# of Interactions    & 8,546   & 649  & 100        &  6,375  & 2,820  & 100    \\ 
\# of Turns           & 29,438  & 2,231  & 1,417   & 22,134  & 9,535  & 1,564    \\ 
Avg. \# of Turns      & 3.44    & 3.44   & 14.17  & 3.47   & 3.38   & 15.64     \\ 
Max \# of Turns       & 9       & 9    & 25       &   8  & 8    & 27     \\ \hline
\end{tabular}
\caption{\label{tab:split_stat}
Statistics for Random and Compositional splits. Test$_{L}$ is a test set with longer interactions, designed to evaluate model's compositional generalization as detailed in Section \ref{sec:exp2}.}
\end{table*}

\section{Experimental Setup}

\subsection{Task}
The objective of the model is to generate a SQL query given the interaction history and current question. We experiment with two versions of interaction history: one that includes only the previous questions, denoted as \textit{QQ}, and another that includes both the previous questions and their corresponding SQL queries, denoted as \textit{QS}. For the \textit{QS} setting, the model is trained with the interaction history using the ground-truth queries, while during the inference phase, the model's own predictions are used in order to simulate a real-world application.

\subsection{Baselines}
\subsubsection{Fine-tuning Models}
We employ both a fine-tuning approach and an in-context learning approach for our baselines. For the fine-tuning approach, we use the T5 models \cite{raffel2020t5}, the general-purpose sequence-to-sequence models as our baseline models. 
We did not use the state-of-the-art models~\cite{cai2022star, xiao2022cqr, qi2022rasat, zheng2022hie, scholak2021picard, li2021pay} for SParC or CoSQL due to their SQL grammar being confined to Spider or standard SQL which is not compatible with our dataset.

\subsubsection{In-context Learning Models}
We use ChatGPT, LLaMA-7B~\cite{touvron2023llama}, and Code-LLaMA-7B~\cite{roziere2023code} for our in-context learning models. 
Instead of a zero-shot approach, we employ few-shot prompting.
We retrieve similar exemplars for each test instance using the BM25 algorithm to use as prompt.
For every experiment with in-context learning setting, we use 20-shot.
For EHRSQL, which is single-turn, computing the similarity for each question is sufficient.
However, for EHR-SeqSQL, which is multi-turn and context-dependent, it's necessary to consider both the current question and the interaction history. Therefore, we use 10 retrieved examples related to the current question and another 10 related to the entire interaction history.
More details such as prompt and few-shot retrieval in our in-context learning baseline are in Appendix \ref{app:llm}. 

\subsection{Evaluation}
The two commonly used metrics in text-to-SQL are Exact Match Accuracy (EM) and Execution Accuracy (EX). However, the EM metric can sometimes be overly strict because it doesn't take into account predictions that have different SQL syntax but yield the same execution result as the ground-truth query.
Thus we use the EX score to measure the model performance based on query execution results.
Before execution, the special tokens are replaced with the generated queries or their execution result through post-processing, thus any errors from the referenced turn will be propagated.

Additionally, for the multi-turn setting, we adopt Interaction Match (IM) and Question Match (QM) following \citet{SPARC} on EX.
IM measures the accuracy of the entire interaction while QM measures the accuracy of each turn.
We utilize different metrics based on the experiment purpose.

\section{Experiments}
\label{experiment}
We now empirically demonstrate the benefits of EHR-SeqSQL in terms of two types of compositional generalization (\autoref{sec:exp1}, \autoref{sec:exp2}), as well as the effectiveness of the special tokens (\autoref{sec:exp3}).

\input{tables/exp5.1}

\subsection{Generalization to Unseen Interaction Goals}
\label{sec:exp1}
\paragraph{Compositional Split}

We first aim to evaluate whether training models in a multi-turn setting can lead to the acquisition of compositional generalization abilities.
To explore this aspect, we split our dataset in a compositional manner, namely \textit{compositional split}. 
This split differs from the \textit{random split} in EHRSQL where the distributions of SQL structures in the training and test sets are nearly identical. In contrast, \textit{compositional split} includes unseen SQL structures in the test set, though these structures can be decomposed into smaller parts that are all present in the training data.

More concretely, we first define the terms \textit{compositions} and \textit{components} and use the concepts to automatically split the dataset.
\textit{Compositions} represent SQL templates in EHRSQL, where condition values, SQL functions (i.e. \texttt{SUM}, \texttt{AVG}, etc), time expressions (i.e. subqueries constraining \textit{last year}, \textit{in 2023}, etc) are masked.
\textit{Components} refer to the decomposed SQL template clauses derived from Stage 2, as explained in Section \ref{sec:3.1.2}.
Each composition, which corresponds to an interaction goal, contains a set of components that exist as individual SQL subqueries in the dataset. 
Table~\ref{tab:composition_example} provides concrete examples of components and compositions. We employ a greedy algorithm to split the dataset, similar to \citet{tmcd}. Table \ref{tab:split_stat} provides the statistics of the random and compositional splits. More details are given in Appendix \ref{app:composition}.

\paragraph{Metric}
We only measure IM to compare the performance of a model trained on EHRSQL and a model trained on EHR-SeqSQL.
Given that an interaction in EHR-SeqSQL corresponds to a question in EHRSQL, correctly predicting every question within an interaction is equal to correctly predicting a single question in EHRSQL.

\paragraph{Result}
The experimental results are shown in Table \ref{Exp1}. 
In a random split, all models exhibit strong performance with both EHRSQL and EHR-SeqSQL.
Still, there is a noticeable performance difference between models specifically trained for SQL generation and those that are not, such as LLaMA and Code-LLaMA, which show relatively worse performance.

However, in the compositional split, models generally perform better with the EHR-SeqSQL dataset, which indicates superior generalization ability in multi-turn setting.
T5 models, in particular, show a significant performance increase, ranging from 7.63\%p (from 67.8\% to 75.43\%), to 30.34\%p (from 52.15\% to 82.49\%). 
T5-base finetuned in \textit{QQ} setting has the best performance.
This suggests that a task-specific fine-tuning enables effective extraction of necessary information only from NLQs without being distracted by other contextual factors. 
Interestingly, T5-3B shows lower performance than T5-base, and this performance gap is particularly pronounced in our proposed compositional split. This might be due to the compositional split having significantly less training data and the evaluation data being perceived as out-of-distribution due to its compositional nature.
This result is consistent with previous observations~\cite{kale-rastogi-2020-text, zhang2023dialogstudio, tmcd} which found that smaller models (T5-base or T5-large) can outperform their larger counterpart, T5-3B, outside the general domain or when dealing with out-of-distribution data.


On the other hand, ChatGPT suffers a performance drop in the \textit{QQ} setting, which is suspected to be due to the absence of predicted SQL queries within the prompt, which leads ChatGPT to have less information on target representation.
Still, a multi-turn setting leads to a significant performance increase in the \textit{QS} setting.
It's worth noting that ChatGPT has better compositional generalization ability in a single-turn setting than fine-tuned models.
LLaMA shows very low performance in every setting, indicating it is not well-suited for generating SQL. Code-Llama performs relatively better and demonstrates clearly better performance with our dataset.
Overall, our experiments show that training with EHR-SeqSQL allows the models to generalize well even with unseen questions.

\subsection{Generalization to Longer Interactions}
\label{sec:exp2}
\paragraph{Longer Interaction Generation}
In this section, we aim to explore whether the models trained with EHR-SeqSQL can comprehend longer interactions, reflecting another form of compositional generalization ability.
To test this, we created a new test set (Test$_L$) composed of much longer interactions with multiple follow-up questions.
This setup is intended to replicate real-world scenarios where users ask a series of questions to incrementally explore a topic of interest~\cite{SPARC}.
For each test set of \textit{random split} and \textit{compositional split}, we created 100 longer interaction instances by connecting related interactions. Details about the longer interaction generation process can be found in Appendix \ref{app:LongerInter}.
The statistics for Test$_L$ for each test set can be found in Table \ref{tab:split_stat}. These interactions are, on average, five times longer than those in the training data.
The Test$_L$ with a random split challenges generalization for longer sequences, whereas the Test$_L$ with a compositional split further complicates this by testing generalization to unseen questions, presenting a scenario that is not only more challenging but also more reflective of practical applications.

\paragraph{Metric}
In this experiment, we investigate whether a model trained on the interactions with \textit{three turns} on average can generalize to interactions with an average of \textit{14 turns}. Given this significant increase in complexity, IM metric is deemed excessively strict. Instead, we report QM to measure the ratio of correctly answered turns.
We further measure the Index of the First Failure (IFF), to capture where the model first generates an incorrect response.
The methodology for calculating the IFF score is detailed in Appendix \ref{app:LongerIFF}.

\input{tables/exp5.2}
\paragraph{Result}
The experiment results are presented in Table \ref{exp:longer_interaction}. Similarly with the findings in Section \ref{sec:exp1}, scores from Test$_{L}$ with random split outperform that from Test$_{L}$ with compositional split. This is likely because the model should generalize to the unseen interaction goals as well as longer interactions in Test$_{L}$ with compositional split.
Interestingly, IFF scores from all models are higher than the average length of interactions in the training data. This suggests that the models can learn interaction-level compositionality and generalize to longer turns after being trained with EHR-SeqSQL.
However, for the \textit{QS} setting, T5-base model had a disadvantage in taking long interactions due to its input length constraint, which led to the lowest IFF score.
Contrasting with fine-tuned T5 models, ChatGPT notably excelled in QM and IFF scores, achieving 12.21 in the \textit{QS} setting, near the ideal 15.17, demonstrating its adaptability to longer interactions.

\subsection{Effects of Special Tokens for SQL}
\label{sec:exp3}
In this section, we evaluate the effect of two special tokens--\texttt{prev\_query} and \texttt{prev\_result}. These two tokens allow models to easily reference either the previous query or its execution result and thereby alleviate the decoding burden. Additionally, \texttt{prev\_result} can further reduce the execution overhead by preventing duplicated subqueries from being executed multiple times. We assess the utility of these tokens from two aspects: model performance and query execution efficiency.

\subsubsection{Effects on Model Performance}

\paragraph{Details}
First, we evaluate the impact of the special tokens on model performance. 
We compare a model trained with queries with the original EHR-SeqSQL, which include the special tokens, to another trained with its standard SQL query version. We ensure that all other training factors, such as model architecture and hyperparameters remain consistent between the two models.
To prevent the test set from being too simple, which could potentially undermine the impact of the special tokens, we use a compositional split. 

\paragraph{Result} As shown in Table~\ref{Exp:3.1}, the incorporation of special tokens consistently enhances the performance across all settings, regardless of the model variant or the type of interaction history.
In the \textit{QQ} setting, where the absence of prior query history makes the contextual questions more challenging, the special tokens contribute to a substantial performance increase. This is because those tokens simplify referencing previous questions.
Specifically, at the question level, the use of these special tokens leads to a performance increase of 14.6\%p (from 75.37 to 89.97), and 30\%p (from 52.49 to 82.49) at the interaction level.
In the \text{QS} setting, on the other hand, the reference is more straightforward because the previous SQL information is given. Still, special tokens enhance model performance by reducing the complexity of the target representation.
ChatGPT demonstrates a robust performance even without the special tokens in the \textit{QS} setting.
We speculate this is because ChatGPT is trained on diverse data, which likely includes the standard SQL. Therefore, it might be familiar with standard text-to-SQL tasks, which is also consistent with the recent finding \cite{liu2023comprehensive}.

\begin{table}[t!]
\centering
\begin{adjustbox}{width=\columnwidth,center}
\renewcommand{\arraystretch}{1.2}
\begin{tabular}{cc|cccl}
\hline
\multicolumn{2}{c|}{\multirow{2}{*}{\textbf{Model}}}                      & \multicolumn{2}{c}{\textbf{SQL}}                   & \multicolumn{2}{c}{\textbf{SQL$^\dag$}} \\ \cline{3-6} 
\multicolumn{2}{c|}{}                                                     & \multicolumn{1}{l}{QM↑} & \multicolumn{1}{l}{IM↑}    & \multicolumn{1}{l}{QM↑}   & IM↑     \\ \hline
\multicolumn{1}{c|}{\multirow{2}{*}{T5-Base}} & \textit{QQ}                       & 75.37              &  52.49                      &     \textbf{89.97}                 & \textbf{82.49}  \\
\multicolumn{1}{c|}{}                         & \textit{QS}                      & 82.10               & 64.35                     &      \textbf{86.46}               & \textbf{75.38}  \\ \hline
\multicolumn{1}{c|}{\multirow{2}{*}{ChatGPT}} & \textit{QQ}                       & 72.07                  & 45.14                     &        \textbf{75.15}           &    \textbf{60.50}    \\
\multicolumn{1}{c|}{}                         & \multicolumn{1}{l|}{\textit{QS}} & \multicolumn{1}{l}{88.22}   & \multicolumn{1}{l}{72.55} & \multicolumn{1}{l}{\textbf{88.23}}     & \textbf{75.43}  \\ \hline
\end{tabular}
\end{adjustbox}
\caption{\label{Exp:3.1}
Comparison of model performance trained with the standard SQL queries and queries with our special tokens (SQL$^\dag$). }
\end{table}



\begin{table}[t!]
\small
\renewcommand{\arraystretch}{1.2}
\begin{adjustbox}{width=\columnwidth,center}
\begin{tabular}{c|cc|cc}
\hline
\multirow{2}{*}{\textbf{Patient}} 
& \multicolumn{2}{c|}{\textbf{SQL}} & \multicolumn{2}{c}{\textbf{SQL${^\dag}$}} \\ \cline{2-5} 
& \multicolumn{1}{c|}{Avg↓}   & Med↓  & \multicolumn{1}{c|}{Avg↓}      & Med↓     \\ \hline
\textbf{1k} & \multicolumn{1}{c|}{0.20}     & 0.09 & \multicolumn{1}{c|}{\textbf{0.09}}  & \textbf{0.01}       \\ \hline
\textbf{10k} & \multicolumn{1}{c|}{3.25}  & 1.17 & \multicolumn{1}{c|}{\textbf{1.15}}  &  \textbf{0.15}  \\ \hline
\textbf{46k} & \multicolumn{1}{c|}{31.57}    & 4.60  & \multicolumn{1}{c|}{\textbf{6.00}}  & \textbf{0.52}
\\ \hline
\end{tabular}
\end{adjustbox} 
\caption{\label{Exp:3.2}
Execution time of SQL queries measured in databases of different sizes. Units are deciseconds ($10^{-1}$).
}
\end{table}

\subsubsection{Effects on Execution time}

\paragraph{Details}
We further evaluate the impact of our special tokens with respect to query execution efficiency. 
Specifically, we compare the execution time of the queries from EHR-SeqSQL and that of their standard SQL version. 
The original database consists of the medical records of 1,000 patients and has a size of 95MB. 
Following the DB construction process in \citet{ehrsql}, we designed two additional larger databases, one with the medical records of 10,000 patients and another with all 46,520 patients that MIMIC-III provides. The size of these databases is 921MB and 5.06GB respectively. The results are reported on the queries from the test set in a random split, which covers all types of SQL queries in EHR-SeqSQL\footnote{We excluded the execution time that is nearly zero.}. We only considered the subset of queries that contain the special token, \texttt{prev\_result}.

\paragraph{Result} The result is shown in Table \ref{Exp:3.2}. 
We observed an 18\% decrease in average execution time, in the original database. On a single query basis, the execution time is reduced at most to 99.89\%, where the original SQL query has five nested queries inside.
The effectiveness of the special token tends to increase with the size of the database. In the largest database, the special token yielded an average time reduction of 81.0\%.
Given that real-world databases typically contain huge amounts of data, we anticipate that the special token will yield a significant practical impact in multi-turn text-to-SQL environments.

\section{Conclusion}
We present EHR-SeqSQL, the first and the largest sequential text-to-SQL dataset designed for EHR databases, aimed at improving interactivity, compositionality, and efficiency in text-to-SQL parsing.
Our experiments show that the question decomposition significantly enhances the model's generalization capabilities for unseen interaction goals and extends its applicability to longer, untrained interactions. Moreover, the introduction of novel SQL-like tokens, which reuse query and execution results from prior queries, enhances execution efficiency and improves the model performance.
We believe our dataset will serve as a valuable testbed for assessing contextual and interactive text-to-SQL tasks and bridge the gap between industry needs and academic research.

\section*{Limitations}
Our dataset is designed to exclude ambiguous or unanswerable questions, focusing instead on the primary objectives of evaluating the model's performance in the text-to-SQL task. It includes converting natural language utterances into appropriate SQL queries and assessing the model's capacity for compositional generalization, generating unseen combinations of subqueries observed during training. This design choice allows our dataset to rigorously test the model's performance. However, we recognize that ambiguous or unanswerable questions are also significant in real-world scenarios. Future research could expand into this area, potentially enhancing the model’s robustness.


\section*{Ethics Statement}
Our main concerns are patient privacy and ensuring that the proposed dataset as well as the construction process are both compliant with the privacy regulations.
The MIMIC-III database, which requires credentialed access via PhysioNet, is a database that contains de-identified medical records of intensive care patients. \citet{ehrsql} further de-identified MIMIC-III for EHRSQL by shifting time and values across the database. We use the database constructed from EHRSQL, which carefully prevents the re-identification of medical records from the questions.
We utilized the ChatGPT api to paraphrase our manually generated question templates. Each question template corresponds to the decomposed queries from EHRSQL and would unlikely contain the sensitive medical information.

\section*{Acknowledgements}
This work was supported by Institute of Information \& Communications Technology Planning \& Evaluation (IITP) grant (No.RS-2019-II190075), National Research Foundation of Korea (NRF) grant (NRF-2020H1D3A2A03100945), and the Korea Health Industry Development Institute (KHIDI) grant (No.HR21C0198), funded by the Korea government (MSIT, MOHW).

\bibliography{anthology,custom}
\bibliographystyle{acl_natbib}

\appendix
\onecolumn

\section{Data Construction Details}

\begin{table*}[h!]
\centering
\renewcommand{\arraystretch}{1.3}
\begin{tabular}{llc}
\Xhline{0.4mm} 
\textbf{Category} & \textbf{Example} & \textbf{\# of questions} \\ \Xhline{0.4mm} 
\multicolumn{1}{@{}l}{\textbf{Independent}} & 
\makecell[l]{
Q1. What is the specific item id of the hemoglobin lab test?} &  \makecell{14,005 \\[-0.8ex] \small(44.22\%)} \\ \hline 
\multicolumn{3}{@{}l}{\textbf{Dependent}} \\ \cline{2-3}
Referential & \makecell[l]{
Q1. Display the hospitalization ids of patient 76173.\\
Q2. How many times each drug was prescribed \underline{within \texttt{A1}} since 2101?
} & \makecell{11,560 \\[-0.8ex] \small(36.50\%)}  \\  \cline{2-3}
Filtering &  \makecell[l]{
Q1. What are the calcium, total lab test values tested during \texttt{A1}? \\
Q2. Retrieve \underline{the last tested case in \texttt{A2}}. }  &  \makecell{5,947 \\[-0.8ex]  \small(18.78\%)} \\
\cline{2-3}
Modifying &  \makecell[l]{
Q1. During \texttt{A1}, what was the last measured value of \texttt{A2}? \\
Q2. \underline{What about the first measured case?} } & \makecell{507 \\[-0.8ex] \small(1.16\%)} \\
\Xhline{0.4mm} 
\end{tabular}
\caption{
\label{tab:data_category}
Category of the questions within EHR-SeqSQL.
}
\end{table*}

\subsection{Details in Stage 2 in Section \ref{sec:3.1.2}}
\label{appendix:stage2_exception}
In stage 2, we decompose the subqueries based on the SQL clauses. Decomposed clauses are parsed according to the logical order of execution of an SQL statement - WHERE, ORDER BY, HAVING, and SELECT. However, there are certain cases that are excluded from this decomposition.
First, if a WHERE clause contains any aliases of table or column names without specifying the original name, it is not decomposed for clarity. 
For example, see \texttt{SELECT DISTINCT T1.C1 FROM ( PREV\_QUERY5 ) AS T1 WHERE \textbf{T1.valuenum = 73.0}}. If decomposed, the meaning of \texttt{T1} would be ambiguous since we do not know the meaning of T1, which is not a column name in MIMIC-III. Thus we ensured the conditions or clauses with table aliases are always used with the subquery where the alias is defined.
Second, we do not decompose the clauses for optional data cleansing SQL clauses. For example, see \texttt{SELECT microbiology events.org\_name FROM microbiologyevents WHERE microbiologyevents.spec\_type\_desc = `foot culture' AND \textbf{microbiologyevents.org\_name IS NOT NULL}}. In this SQL query, the bolded subquery is intended to ensure all the selected rows have contents, excluding any NULL values. It is an optional condition and a SQL writing style choice which is not included in the original question, so we choose not to decompose such clauses.
Also, the ten shortest question templates from EHRSQL are maintained without further splitting or modifications, which are not likely to be asked through multiple sentences.
These include questions asking about the drug intake method, the cost of a lab test, or the number of current patients.
We do not decompose \texttt{GROUP BY} but combine it either with \texttt{SELECT} or \texttt{HAVING}, since they are not explicitly expressed in the natural language questions \cite{guo2019towards, wu2021data}.

\subsection{Details in Stage 3 in Section \ref{sec:3.1.3}}
\label{appendix:stage3_details}
The BPE algorithm was applied to merge the subqueries from both Stage 1 and Stage 2. Specifically, we first derived SQL templates by masking condition values in the SQL queries obtained in Stage 2. Each template was treated as a token in the BPE algorithm, and the syntactically mergeable pairs among the most frequent token bigrams were repeatedly merged.
A syntactically mergeable pair refers to a pair of queries that were originally a single SQL query but were separated due to our decomposition strategy. Since BPE algorithm calculates frequency without considering the relationship of token bigrams, we added this constraint to make every resultant SQL query to be executable.
Then, we sampled half of each randomly to maximize the complexity of each subquery and the diversity of each interaction. Considering the total number of turns in each stage, we merged bigrams that appear more than 100 times in the subqueries from Stage 1 and bigrams that appear more than 150 times in the subqueries from Stage 2. The final SQL queries for EHR-SeqSQL are acquired throughout three stages.

\subsection{List of NLQ templates}
\label{app:NLQ}
In Section \ref{sec:3.2}, we mentioned the process for rule-based SQL-to-NLQ generation. Detailed steps of this process can be found in Table \ref{tab:nlq_generation}. We used specific SQL templates paired with NLQ templates for this process, as presented in Table \ref{tab:nlq_template}. Each placeholder in the NLQ template is determined by the DB schema and the values present in the actual SQL query. A few examples showcasing the pairing of DB schema and NL expressions are available in Table \ref{tab:nlq_expression}. The placeholders related to \textit{value}, \textit{operation}, and \textit{time} are used in the same way as described by \citet{ehrsql}.

\input{tables/nlq_generation}
\input{tables/nlq_template}
\input{tables/nlq_placeholder}

\subsection{Prompt for paraphrasing}
\begin{figure*}[b]
\centering
\includegraphics{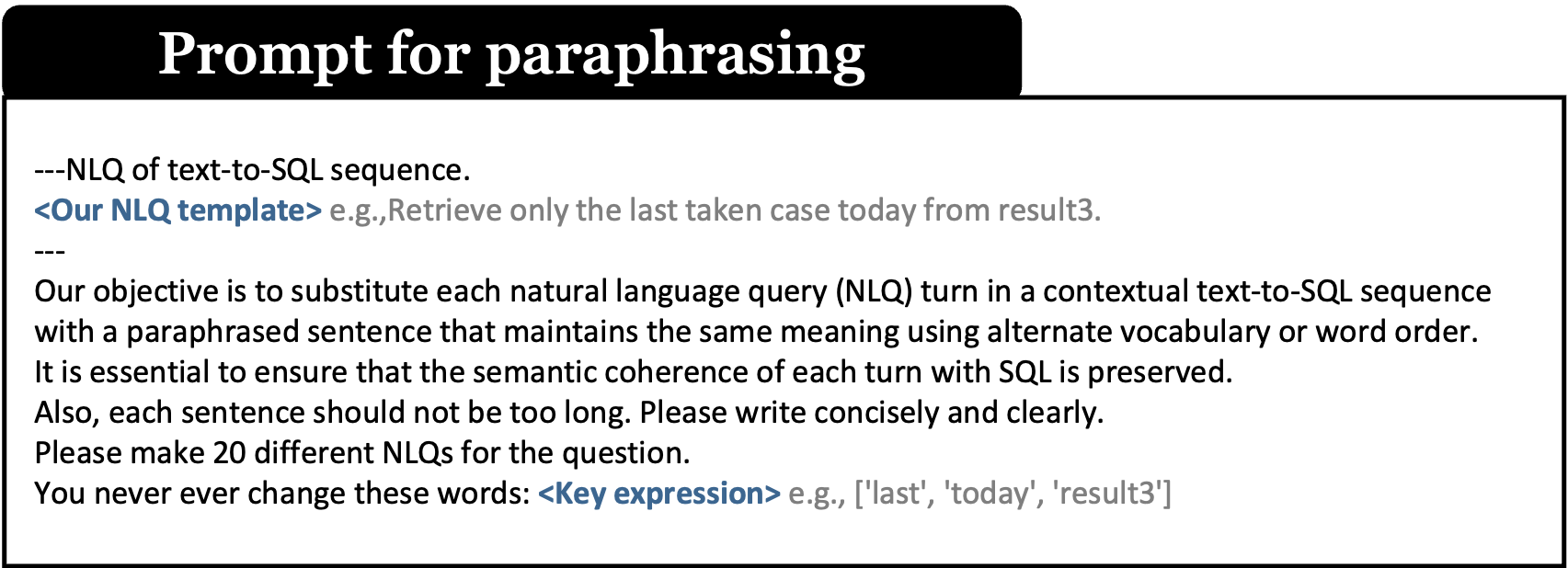}
    \caption{ Prompt for paraphrasing. }
\label{app:para_prompt}
\end{figure*}

Figure \ref{app:para_prompt} shows the prompt for paraphrasing the questions at an interaction level. For each NLQ template, key expressions (such as condition values, reference indices, etc.) that must be strictly preserved were indicated in order to maintain consistency. 
During the data construction process, we utilized the ChatGPT API to paraphrase template questions. Prior to paraphrasing, all specific values within each template are replaced with representative, generic values for their respective slots. Once paraphrased, these generic values are then realized back to their original form. This process naturally prevents any sensitive information from being sent to the ChatGPT server.

\section{ Details on the Compositional Split }

\label{app:composition}
\input{tables/compositions.tex}

The concept of compositions and components is based on the SQL query since it is the common factor across an EHRSQL instance and its corresponding EHR-SeqSQL instance. 
Table \ref{tab:composition_example} provides three examples of composition and components. 
There are EHRSQL questions which is an interaction goal of the corresponding interaction in EHR-SeqSQL, and their templates where the same anonymizing logic in deriving composition is applied.

We use a greedy algorithm to automatically split our data into training and test sets. Starting with all compositions assigned to the training set, we iteratively allocate a composition that has the maximum number of unique components to the test set while constraining all the components in the composition to exist in the training set, until no composition can be further assigned to the test set.
You can find that the components in the last row are all present in the components of the first two rows. Thus, according to our split algorithm, if both the first two compositions are in the training set, the last composition can be assigned to the test set.

\section{ Longer Interaction }
\subsection{Generation Process}
\label{app:LongerInter}

\begin{figure}
\centering
    \includegraphics[width=12cm,height=5.3cm]{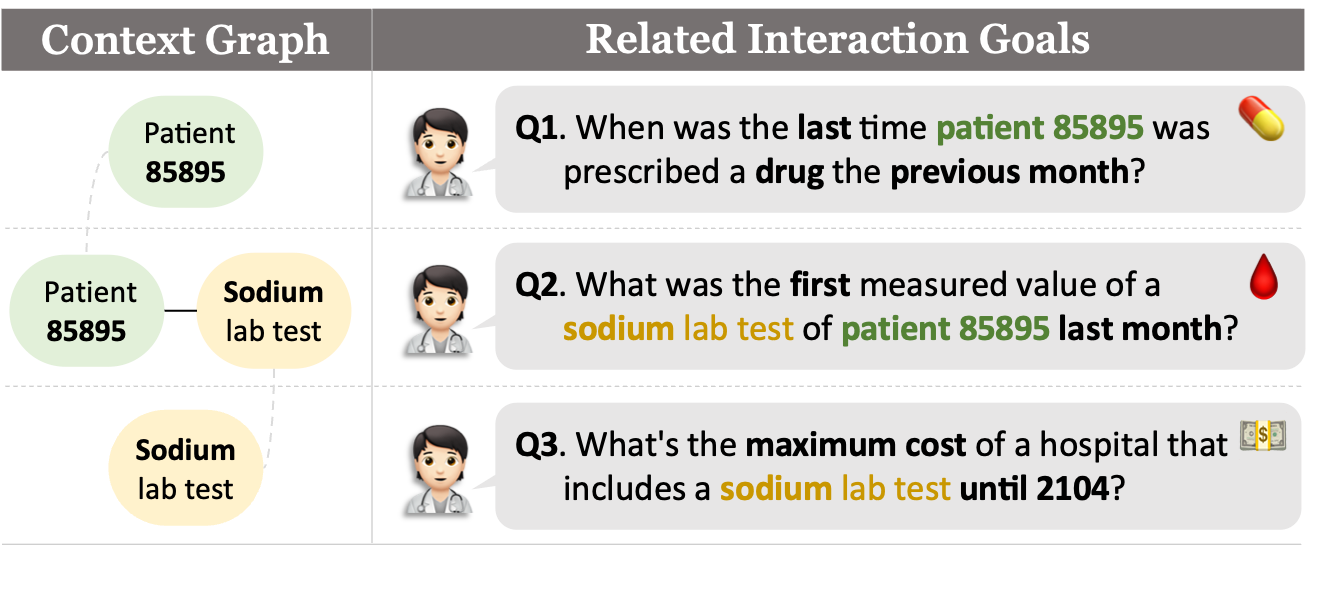}
    \caption{Related interaction goals and their context graph. }
    \label{fig:longer_sequence}
\end{figure}

\begin{figure*}
\centering
\includegraphics[width=15.5cm,height=23cm]{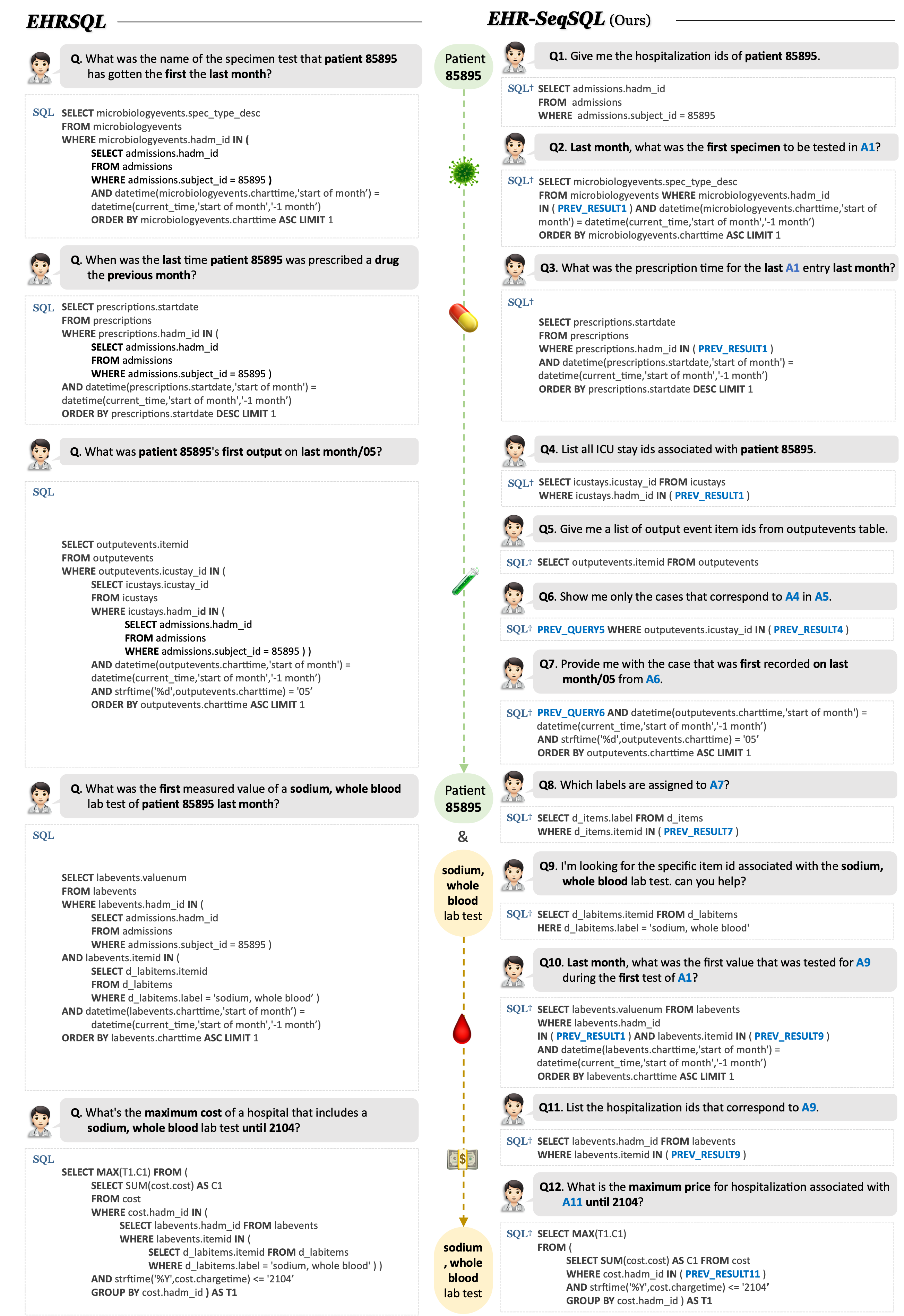}
    \caption{ Example of Longer Interaction. }
\label{app:longer_interaction}
\end{figure*} 

As given in Figure \ref{fig:longer_sequence}, we define the concept of related interaction goal by using the context graph. Each interaction goal has its own context graph, whose nodes are defined by the specific condition values (e.g. conditions for patient, drug, or lab tests) in the original EHRSQL. 
Two questions are deemed related if they have overlapping condition values. 
Q1 and Q2 are related because they share the same patient 85895. However, Q1 and Q3 are not directly related without Q2 as a bridge.
While we used three independent questions from the EHRSQL to demonstrate this concept in Figure \ref{fig:longer_sequence}, a longer and context-dependent interaction used in Section \ref{sec:exp2} is depicted in Figure \ref{app:longer_interaction}. In this example for longer interaction, five EHRSQL questions are connected, each paired with an original SQL query. In contrast, EHR-SeqSQL involves context-dependent interaction spanning twelve turns, each paired with a query with our special tokens for SQL.

\subsection{IFF Score Calculation}
\label{app:LongerIFF}

\begin{center}
    $IFF = \begin{cases} 
    n + 1, & \text{if all turns are correct} \\
    k, & \text{otherwise}
    \end{cases}$
\end{center}

$k$ denotes the specific turn number in the interaction where the first incorrect response occurs. $n$ represents the total number of turns in the interaction.
The final IFF score for a test set is calculated as the average IFF score across all interactions.
Hence, if the model achieves a perfect score for every interaction, the IFF score becomes one plus the average number of turns of all interactions.

Note that the perfect IFF score for QQ and QS setting is different in Table \ref{exp:longer_interaction}. This is because we removed the test sample that exceeds the maximum token length of T5 when the interaction history is concatenated with the current question for a fair comparison.



\section{Fine-tuning Baseline}
\label{app:ft}
\subsection{Configuration}
We fine-tuned the T5-Base and T5-3B using the Adam optimizer, with a global batch size of 32. The learning rate were set to 1e-4 for T5-Base and 5e-5 for T5-3B, respectively. For the other hypterparameter configurations, we followed the settings used in EHRSQL.
All experiments were carried out on either a single A100 80G GPU or a A6000 48G GPU. The training process typically took around 10 hours for T5-Base and around 24 hours for T5-3B.


\section{In-context Learning Baseline}
\label{app:llm}

\subsection{Prompt Configuration}
\label{app:prompt}
\begin{figure*}
\centering
\includegraphics[width=16cm,height=11cm]{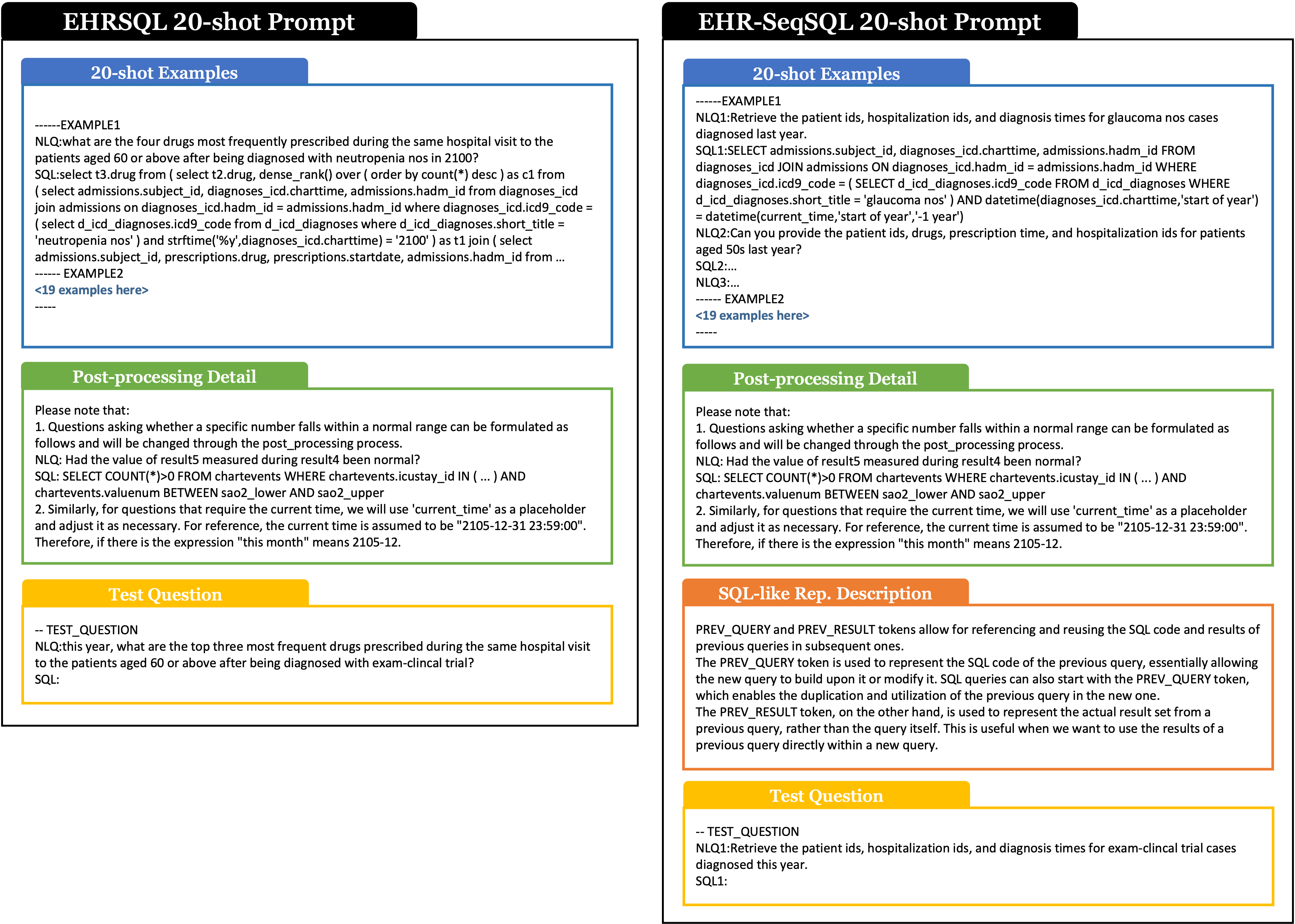}
    \caption{ Prompt configuration for ChatGPT. }
\label{app:llm_prompt}
\end{figure*}

Figure \ref{app:llm_prompt} demonstrates how we configured the prompt of ChatGPT for both the EHRSQL and EHR-SeqSQL.
Unlike EHRSQL where the target representation is standard SQL, EHR-SeqSQL include the special tokens, which are first introduced in this work. Thus, we include a simple description for \texttt{PREV\_QUERY} and \texttt{PREV\_RESULT} tokens in the prompt.
In the \textit{QS} setting during the inference process, for the current turn, the interaction history includes both previous questions and queries generated by ChatGPT.

\subsection{ Few-Shot Learning Approach for EHR-SeqSQL }
\label{app:bm25}
\begin{figure*}[b]
\centering
\includegraphics[width=16cm]{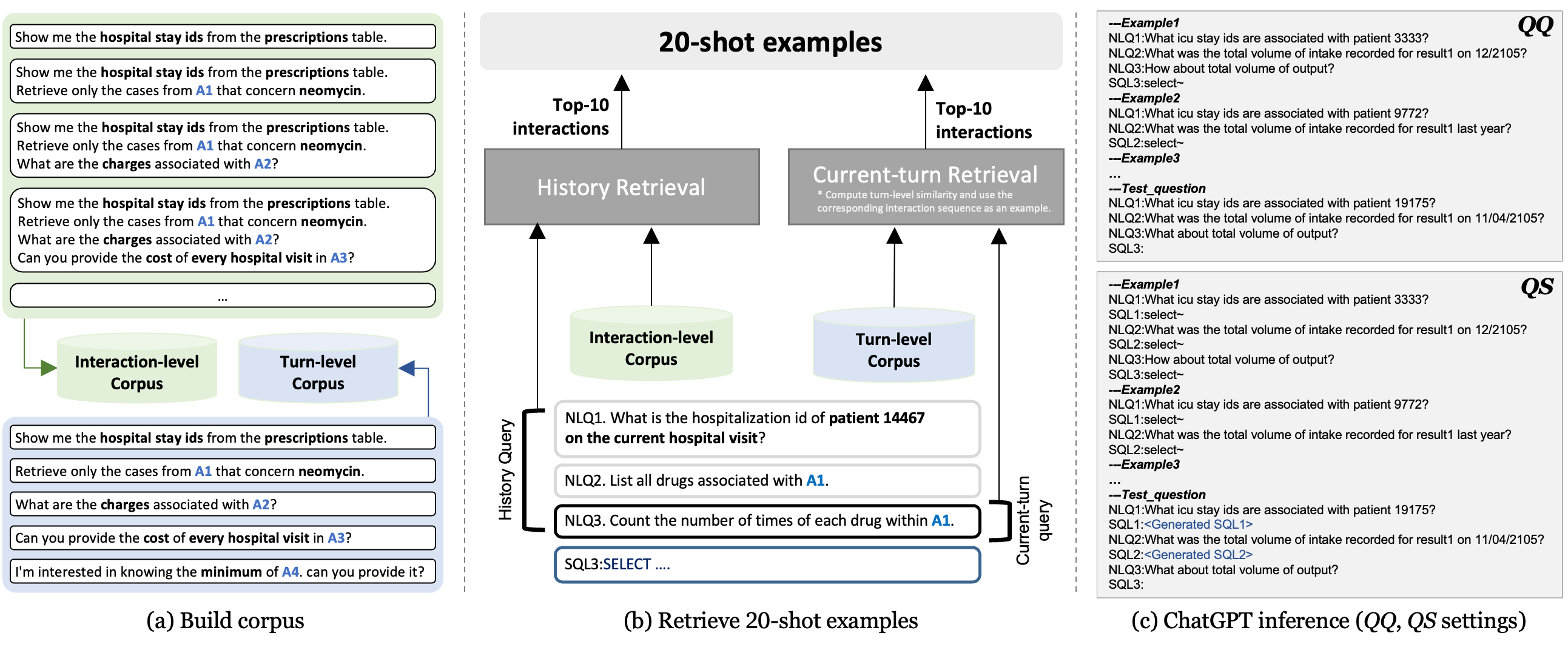}
    \caption{ Overview of ChatGPT baseline. In (a), we built two corpora using training data. For turn-level, each question is stored independently, while in an interaction-level, questions accumulate within each interaction.
    (b) shows the retrieval process of the top 20 examples based on BM25 similarity: 10 from using the entire interaction as a query (history retrieval) and another 10 from using just the current turn (current-turn retrieval). The interaction-level and turn-level corpus from (a) are used respectively.
    (c) deals with prompt based on different versions of interaction history. In the \textit{QQ} setting, only 20 target representations are given, whereas in the \textit{QS} setting, more target representations are shown, depending on the length of the example interactions.
    }
\label{app:chatGPT_overview}
\end{figure*}

We developed a new method for few-shot learning for the multi-turn, context-dependent setting of EHR-SeqSQL. This method is detailed in Figure \ref{app:chatGPT_overview}. To begin, we created two corpora: one for interaction-level training data and another for turn-level training data. For our few-shot learning approach, we retrieved examples, with half based on the interaction history and the other half based on the current question. In our experiments, we used a total of 20 examples for few-shot learning.

\section{Use Cases of the Special Tokens}
Table \ref{tab:exec_time_example} illustrates an example of the interactions in EHR-SeqSQL where the special token significantly reduce the length of target representation as well as the execution time. Due to readability and spatial issues, the standard SQL version of each turn has been omitted from the table. You can see the standard SQL would be quite lengthy by looking at the SQL$^\dag$ that the used special token refers to.

\input{tables/reduced_exec_time}

\end{document}

%% file: tables/exp5.1.tex



\begin{table*}
\centering
\renewcommand{\arraystretch}{1.1}

\begin{tabular}{c|ccc|ccc}
\hline
\multirow{3}{*}{\textbf{Model}} & \multicolumn{3}{c}{\textbf{Random}}                         & \multicolumn{3}{|c}{\textbf{Compositional}}                  \\ \cline{2-7} 
                       & \multirow{2}{*}{\textbf{EHRSQL}} & \multicolumn{2}{c|}{\textbf{EHR-SeqSQL}} & \multirow{2}{*}{\textbf{EHRSQL}} & \multicolumn{2}{c}{\textbf{EHR-SeqSQL}} \\ \cline{3-4} \cline{6-7} 
                       &                         & \textit{QQ}        & \textit{QS}     &                         & \textit{QQ}       &\textit{QS}      \\ \hline
T5-Base                & $94.25_{\pm0.54}$         & $94.69_{\pm0.01}$    & $95.38_{\pm1.48}$       & $52.15_{\pm0.85}$           & $82.49_{_\pm1.80}$   &  $75.38_{\pm1.94}$     \\
T5-3B                  & $90.68_{\pm9.47}$         & $93.07_{\pm3.71}$      & $92.38_{\pm4.9}$       &       $54.90_{\pm3.51}$      & $71.39_{\pm1.05}$ &  $70.07_{\pm1.85}$    \\
ChatGPT                & 91.22                     &  80.74                 & 91.37       &    67.80      &    60.50   & 75.43            \\ 


LLaMA-7B                & 59.32                     &  61.79                 & 41.60       &    17.34      &    12.73   & 26.42            \\ 
Code-LLaMA-7B                & 64.56                     &  76.58                 & 70.72       &    32.30      &    48.62   & 47.30            \\ \hline
\end{tabular}
\caption{\label{Exp1}
Model performances in two different splits. 
For the fine-tuning models, we trained each model with three different random seeds. We report the average score and the standard deviation.
}

\end{table*}

%% file: tables/exp5.2.tex
\begin{table}[t!]
\centering
\begin{adjustbox}{width=\columnwidth,center}
\begin{tabular}{cc|cc|cl}
\hline
\multicolumn{2}{c|}{\multirow{2}{*}{\textbf{Model}}}                      & \multicolumn{2}{c|}{\textbf{{Random}}}                   & \multicolumn{2}{c}{\textbf{Compositional}} \\ \cline{3-6} 
\multicolumn{2}{c|}{}                                                     & QM↑ &  IFF↑    &  QM↑   & IFF↑     \\ \hline
\multicolumn{1}{c|}{\multirow{2}{*}{T5-Base}} & \textit{QQ}                       & 54.13             &  \makecell{5.70 \\[-0.8ex] \small (15.17)}                      &     50.58       & \makecell{5.08 \\[-0.8ex] \small (16.64)}  \\
\multicolumn{1}{c|}{}                         & \textit{QS}                      &  71.33               & \makecell{4.51 \\[-0.8ex] \small (8.03)}                    &     61.58         & \makecell{3.72 \\[-0.8ex] \small (8.31)}  \\ \hline
\multicolumn{1}{c|}{\multirow{2}{*}{ChatGPT}} & \textit{QQ}                       & 92.94                  & \makecell{11.24 \\[-0.8ex] \small(15.17)}                     &        73.66        &    \makecell{5.98 \\[-0.8ex] \small(16.64)}    \\
\multicolumn{1}{c|}{}                         & \multicolumn{1}{l|}{\textit{QS}} & \textbf{94.57}   & \makecell{\textbf{12.21} \\[-0.8ex] \small(15.17)} &   \textbf{79.67} & \makecell{\textbf{6.81} \\[-0.8ex] \small(16.64)}  \\ \hline
\end{tabular}
\end{adjustbox}
\caption{\label{exp:longer_interaction}
Model performances on longer interactions. For the IFF score, the perfect scores in parentheses.}
\end{table}

%% file: tables/nlq_generation.tex
\begin{table*}[]
\centering
\resizebox{\textwidth}{!}{\begin{tabular}{p{0.05in}|p{2in}|p{2in}|p{2in}|p{2in}}
\hline
\multicolumn{1}{c|}{\textbf{\begin{tabular}[c]{@{}c@{}} interaction\\      goal\end{tabular}}} &
  \multicolumn{4}{c}{\begin{tabular}[c]{@{}c@{}}What were the top four   frequent drugs that patients were prescribed \\      within the same month after having been prescribed with nateglinide last   year?\end{tabular}} \\ \hline
\textbf{idx} &
  \textit{\textbf{SQL Query}} &
  \textit{\textbf{SQL Template}} &
  \textit{\textbf{NLQ Template}} &
  \textit{\textbf{Generated NLQ}} \\ \hline
1 &
  \texttt{SELECT   admissions.subject\_id, prescriptions.startdate, FROM prescriptions JOIN   admissions ON prescriptions.hadm\_id = admissions.hadm\_id WHERE   prescriptions.drug = 'nateglinide' {[}time\_filter\_global1{]}} &
  \texttt{SELECT   table.column, table.column   FROM table JOIN table ON table.column = table.column WHERE table.column =   {[}val\_placeholder{]} {[}time\_filter\_global1{]}} &
  List   all {[}SELECT.col:admission.subject\_id{]} and their {[}SELECT.col:prescriptions.startdate{]} associated with {[}val\_placeholder:nateglinide{]}   {[}time\_filter\_global1:last year{]}. &
  List   all patient ids and their prescription time associated with nateglinide last year. \\ \hline
2 &
  \texttt{SELECT   admissions.subject\_id, prescriptions.drug, prescriptions.startdate, FROM   prescriptions JOIN admissions ON prescriptions.hadm\_id = admissions.hadm\_id   WHERE {[}time\_filter\_global1{]}} &
  \texttt{SELECT table.column, table.column, table.column FROM table   JOIN table ON table.column = table.column WHERE {[}time\_filter\_global1{]}} &
  List   all {[}SELECT.col:admission.subject\_id{]} and their {[}SELECT.col:prescriptions.drug{]}   and {[}SELECT.col:prescriptions.startdate{]}   {[}time\_filter\_global1:last year{]}. &
  List   all patient ids and their drugs and prescription time last year. \\ \hline
3 &
  \texttt{SELECT   T2.drug, DENSE\_RANK() OVER ( ORDER BY COUNT(*) DESC ) AS C1 FROM (   {[}PREV\_QUERY1{]} ) AS T1 JOIN ( {[}PREV\_QUERY2{]} ) AS T2 ON T1.subject\_id =   T2.subject\_id WHERE T1.startdate \textless T2.startdate {[}time\_filter\_within{]}   GROUP BY T2.drug} &
  \texttt{SELECT   table.column, DENSE\_RANK()   OVER ( ORDER BY COUNT(*) DESC ) AS column FROM ( {[}PREV1{]} ) AS table JOIN ( {[}PREV2{]} ) AS table   ON table.column = table.column WHERE table.column \textless table.column   {[}time\_filter\_within{]} GROUP BY table.column} &
  List   the frequency rankings of {[}PREV:2{]} that patients received {[}time\_filter\_within:within   the same month{]} after the {[}PREV:1{]}. &
  List   the frequency rankings of A2   that patients received within the same month after A1. \\ \hline
4 &
  \texttt{SELECT   T3.drug FROM ( {[}PREV\_QUERY3{]} ) AS T3 WHERE T3.C1 \textless{}=   {[}\texttt{n\_rank}{]}} &
  \texttt{SELECT table.column FROM ( {[}PREV3{]} ) AS table WHERE   table.column \textless{}= {[}n\_rank{]}} &
  List   the top {[}n\_rank:four{]} {[}SELECT.col:prescriptions.drug{]} in {[}PREV:3{]}. &
  List the top four drugs in A3. \\ \hline
\end{tabular}}
\caption{SQL-to-NLQ Generation Process.}
\label{tab:nlq_generation}
\end{table*}

%% file: tables/nlq_template.tex
\begin{table*}
  \centering 
  \renewcommand{\arraystretch}{1.3}
  \resizebox{1\textwidth}{!}{%
  \begin{tabular}{ll}
    \hline
    \large{\textbf{SQL template}} & \large{\textbf{NLQ template} }
    \\ \hline
    SELECT ( [PREV0] ) - ( [PREV1] ) & What is the difference between the [PREV0] and [PREV1]? \\ \hline
        SELECT ( [PREV0] ) [comparison] ( [PREV1] ) & Is [PREV0] [comparison] than [PREV1]? \\ \hline
        SELECT [agg\_function](table.column) FROM ( [PREV] ) AS table & What is the [SELECT.col] of [PREV]? \\ \hline
        SELECT COUNT( DISTINCT table.column ) FROM ( [PREV] ) AS table & Count the number of patients in [PREV]. \\ \hline
        SELECT COUNT( DISTINCT table.column ) FROM table \#cond\_parsed & Count the number of [PREV-1]. \\ \hline
        SELECT COUNT( DISTINCT table.column ) FROM table WHERE [time\_filter\_global1] & Count the number of [PREV-1]. \\ \hline
        SELECT COUNT(*) FROM table WHERE table.column = ( [PREV] ) & Count the number of [SELECT.col] associated with [PREV]. \\ \hline
        SELECT COUNT(*) FROM table WHERE table.column = [val\_placeholder] & Count the number of [val\_placeholder]. \\ \hline
        SELECT COUNT(*) FROM table WHERE table.column IN ( [PREV] ) & Count the number of [SELECT.col] associated with [PREV]. \\ \hline
        SELECT COUNT(*)>0 FROM table WHERE table.column = [val\_placeholder] & Has [val\_placeholder] been admitted to the hospital? \\ \hline
        SELECT COUNT(*)>0 FROM table WHERE table.column IN ( [PREV] ) & Are there any [SELECT.col] in [PREV]? \\ \hline
        SELECT SUM(table.column) FROM table WHERE table.column IN ( [PREV] ) & What is the [SELECT.col] associated with [PREV]? \\ \hline
        SELECT SUM(table.column) FROM table WHERE table.column IN ( [PREV] )  & What is the total amount of [PREV-1]? \\ \hline
        SELECT table.column FROM ( [PREV] ) AS table WHERE table.column [n\_times] & Which [SELECT.col] is [n\_times] in [PREV]? \\ \hline
        SELECT table.column FROM ( [PREV] ) AS table WHERE table.column <= [n\_rank] & List top [n\_rank] [SELECT.col] in [PREV]. \\ \hline
        SELECT table.column FROM table & List all [SELECT.col] from [FROM.table]. \\ \hline
        SELECT table.column FROM table WHERE [age\_group] & List all [SELECT.col] associated with patients aged [age\_group]. \\ \hline
        SELECT table.column FROM table WHERE table.column = ( [PREV] ) & What is the [SELECT.col] of [PREV]? \\ \hline
        SELECT table.column FROM table WHERE table.column = [val\_placeholder] & List all [SELECT.col] of [val\_placeholder]. \\ \hline
        SELECT table.column FROM table WHERE table.column IN ( [PREV] ) & List all [SELECT.col] associated with [PREV]. \\ \hline
        SELECT table.column, table.column FROM table & List all [SELECT.col.0] and [SELECT.col.1]. \\ \hline
        [PREV] [time\_filter\_exact1] & What was the [time\_filter\_exact2] measured case from [PREV-1]? \\ \hline
        [PREV] [time\_filter\_global1\_dec1] & Retrieve only the cases [time.verb] [time\_filter\_global1\_dec1] from [PREV]. \\ \hline
        [PREV] [time\_filter\_global1\_dec2] & Retrieve only the cases [time.verb] [time\_filter\_global1\_dec2] from [PREV]. \\ \hline
        [PREV] [time\_filter\_global1] & Retrieve only the cases [time.verb] [time\_filter\_global1] from [PREV]. \\ \hline
        [PREV] AND [age\_group] & Retrieve only the cases associated with patients aged [age\_group] from [PREV\_QEURY]. \\ \hline
        [PREV] AND table.column = ( [PREV] ) & Retrieve only the cases associated with [PREV] from [PREV]. \\ \hline
        [PREV] AND table.column IN ( [PREV] ) & Retrieve only the cases associated with [PREV] from [PREV]. \\ \hline
        [PREV] AND table.column IS NULL & What is the current one in [PREV]? \\ \hline
        [PREV] WHERE [age\_group] & Retrieve only the cases associated with patients aged [age\_group] from [PREV]. \\ \hline
        [PREV] WHERE [time\_filter\_global1] & Retrieve only the cases [time.verb] [time\_filter\_global1] from [PREV]. \\ \hline
        [PREV] WHERE table.column = ( [PREV] ) & Retrieve only the cases associated with [PREV] from [PREV]. \\ \hline
        [PREV] WHERE table.column IN ( [PREV] ) & Retrieve only the cases associated with [PREV] from [PREV\_QEURY]. \\ \hline
  \end{tabular}%
  }
  \caption{SQL \& NLQ template.}
  \label{tab:nlq_template}
\end{table*}

%% file: tables/nlq_placeholder.tex
\begin{table*}[]
\centering
\small
\begin{tabular}{@{}l|l@{}}
\hline
\renewcommand{\arraystretch}{1.3}
\textbf{DB schema}                  & \textbf{NL expression}\\ \hline
admissions.admittime                & admission time       \\ \hline
admissions.dob                      & date of birth        \\ \hline
admissions.dod                      & date of death        \\ \hline
admissions.subject\_id              & patient              \\ \hline
chartevents.charttime               & chart time           \\ \hline
chartevents.itemid                  & vital sign item id   \\ \hline
chartevents.valuenum                & value of vital sign  \\ \hline
cost.hadm\_id                       & hospital stay        \\ \hline
diagnoses\_icd.charttime            & time of diagnosis    \\ \hline
diagnoses\_icd.icd9\_code           & diagnosis ICD-9 code \\ \hline
inputevents\_cv.amount              & volume of intake     \\ \hline
inputevents\_cv.itemid              & input event item id  \\ \hline
labevents.itemid                    & lab test             \\ \hline
labevents.itemid                    & lab test item id     \\ \hline
labevents.valuenum                  & value of lab test    \\ \hline
microbiologyevents.org\_name        & organism name        \\ \hline
microbiologyevents.spec\_type\_desc & microbiology test    \\ \hline
outputevents.itemid                 & output event item id \\ \hline
prescriptions.drug                  & drug                 \\ \hline
prescriptions.startdate             & prescription time    \\ \hline
procedures\_icd.charttime           & time of procedure    \\ \hline
procedures\_icd.hadm\_id            & hospital stay        \\ \hline
procedures\_icd.icd9\_code          & procedure ICD-9 code \\ \hline
procedures.icd9\_code               & procedure            \\ \hline
\end{tabular}
  \caption{Examples of DB schema \& NL expression pairs.}
  \label{tab:nlq_expression}
\end{table*}

%% file: tables/compositions.tex
\begin{table*}
  \small 
  \renewcommand{\arraystretch}{1.3}
  \resizebox{1\textwidth}{!}{%
  \begin{tabular}{p{1in} p{1in} p{2.5in} p{3in}}
    \hline
    \textbf{Interaction Goal} & \textbf{Template} & \textbf{Composition} &  \textbf{Set of Components} \\ \hline
    
    Is the value of \textcolor{magenta}{glucose} of patient \textcolor{teal}{71192} \textcolor{blue}{last} measured \textcolor{orange}{on the first hospital visit} less than the \textcolor{blue}{second to last} value measured \textcolor{orange}{on the first hospital visit}?
    & Is the value of \textcolor{magenta}{\{lab\_name\}} of patient \textcolor{teal}{\{patient\_id\}} \textcolor{blue}{[time\_filter\_exact2]} measured \textcolor{orange}{[time\_filter\_global2]} \textcolor{purple}{[comparison]} than the \textcolor{blue}{[time\_filter\_exact1]} value measured \textcolor{orange}{[time\_filter\_global1]}?       & \texttt{SELECT ( SELECT labevents.valuenum FROM labevents WHERE labevents.hadm\_id IN ( SELECT admissions.hadm\_id FROM admissions WHERE admissions.subject\_id = \textcolor{teal}{\{patient\_id\}} \textcolor{orange}{[time\_filter\_global1]} ) AND labevents.itemid IN ( SELECT d\_labitems.itemid FROM d\_labitems WHERE d\_labitems.label = \textcolor{magenta}{\{lab\_name\}} ) \textcolor{blue}{[time\_filter\_exact1]} ) \textcolor{purple}{<} ( SELECT labevents.valuenum FROM labevents WHERE labevents.hadm\_id in ( SELECT admissions.hadm\_id FROM admissions WHERE admissions.subject\_id = \textcolor{teal}{\{patient\_id\}} \textcolor{orange}{[time\_filter\_global2]} ) AND labevents.itemid IN ( SELECT d\_labitems.itemid FROM d\_labitems WHERE d\_labitems.label = \textcolor{magenta}{\{lab\_name\}} ) \textcolor{blue}{[time\_filter\_exact2]} )}   
    & \begin{tabular}[t]{@{}p{3in}@{}}
    1. \texttt{SELECT admissions.hadm\_id FROM admissions} \\
    2. \texttt{[PREV\_QUERY] WHERE admissions.subject\_id = \textcolor{teal}{\{patient\_id\}}} \\
    3. \texttt{[PREV\_QUERY] AND \textcolor{orange}{[time\_filter\_global1]}} \\
    4. \texttt{SELECT d\_labitems.itemid FROM d\_labitems} \\
    5. \texttt{[PREV\_QUERY] WHERE d\_labitems.label = \textcolor{magenta}{\{lab\_name\}}} \\
    6. \texttt{SELECT labevents.valuenum FROM labevents} \\
    7. \texttt{[PREV\_QUERY] WHERE labevents.hadm\_id IN ( [PREV\_RESULT] )} \\
    8. \texttt{[PREV\_QUERY] AND labevents.itemid IN ( [PREV\_RESULT] )} \\
    9. \texttt{[PREV\_QUERY] \textcolor{blue}{[time\_filter\_exact1]}} \\
    10. \texttt{[PREV\_QUERY] \textcolor{blue}{[time\_filter\_exact2]}} \\
    11. \texttt{SELECT ( [PREV\_RESULT] ) \textcolor{purple}{[comparison]} ( [PREV\_RESULT] )} \end{tabular} \\ \hline
    
    What was the \textcolor{purple}{maximum} \textcolor{magenta}{arterial bp [diastolic]} of patient \textcolor{teal}{18866} \textcolor{orange}{yesterday}?
    & What was the \textcolor{purple}{[agg\_function]} \textcolor{magenta}{\{vital\_name\}} of patient \textcolor{teal}{{patient\_id}} \textcolor{orange}{[time\_filter\_global1]}?
    & \texttt{SELECT \textcolor{purple}{[agg\_function]}(chartevents.valuenum) FROM chartevents WHERE chartevents.icustay\_id IN ( SELECT icustays.icustay\_id FROM icustays WHERE icustays.hadm\_id IN ( SELECT admissions.hadm\_id FROM admissions WHERE admissions.subject\_id = \textcolor{teal}{{patient\_id}} ) ) AND chartevents.itemid IN ( SELECT d\_items.itemid FROM d\_items WHERE d\_items.label = \textcolor{magenta}{\{vital\_name\}} AND d\_items.linksto = 'chartevents' ) \textcolor{orange}{[time\_filter\_global1]}}
    & \begin{tabular}[t]{@{}p{3in}@{}}
     1. \texttt{SELECT admissions.hadm\_id FROM admissions} \\
     2. \texttt{[PREV\_QUERY] WHERE admissions.subject\_id = \textcolor{teal}{\{patient\_id\}}} \\
    3. \texttt{SELECT icustays.icustay\_id FROM icustays }\\
    4. \texttt{[PREV\_QUERY] WHERE icustays.hadm\_id IN ( [PREV\_RESULT] )} \\
    5. \texttt{SELECT d\_items.itemid FROM d\_items }\\
    6. \texttt{[PREV\_QUERY] WHERE d\_items.label = \textcolor{magenta}{\{vital\_name\}} AND d\_items.linksto = 'chartevents'} \\
    7. \texttt{SELECT chartevents.valuenum FROM chartevents} \\
    8. \texttt{[PREV\_QUERY] WHERE chartevents.icustay\_id IN ( [PREV\_RESULT] )} \\
    9. \texttt{[PREV\_QUERY] AND chartevents.itemid IN ( [PREV\_RESULT] )} \\
    10. \texttt{[PREV\_QUERY] \textcolor{orange}{[time\_filter\_global1]}} \\
    11. \texttt{SELECT \textcolor{purple}{[agg\_function]}(chartevents.valuenum) FROM chartevents WHERE chartevents.icustay\_id IN ( [PREV\_RESULT] ) AND chartevents.itemid IN ( [PREV\_RESULT] ) \textcolor{orange}{[time\_filter\_global1]}} \end{tabular} \\ \hline
     Is the \textcolor{magenta}{arterial bp [diastolic]} of patient \textcolor{teal}{25461} \textcolor{blue}{last} measured \textcolor{orange}{on the last icu visit} greater than \textcolor{blue}{the second to last} value measured \textcolor{orange}{on the last icu visit}?
     & Is the \textcolor{magenta}{\{vital\_name\}} of patient \textcolor{teal}{\{patient\_id\}} \textcolor{blue}{[time\_filter\_exact2]} measured \textcolor{orange}{[time\_filter\_global2]} \textcolor{purple}{[comparison]} than the \textcolor{blue}{[time\_filter\_exact1]} value measured \textcolor{orange}{[time\_filter\_global1]}?
     & \texttt{SELECT ( SELECT chartevents.valuenum from chartevents WHERE chartevents.icustay\_id IN ( SELECT icustays.icustay\_id from icustays WHERE icustays.hadm\_id IN ( SELECT admissions.hadm\_id from admissions WHERE admissions.subject\_id = \textcolor{teal}{\{patient\_id\}} ) \textcolor{orange}{[time\_filter\_global1]} ) AND chartevents.itemid IN ( SELECT d\_items.itemid from d\_items WHERE d\_items.label = \textcolor{magenta}{\{vital\_name\}} AND d\_items.linksto = `chartevents' ) \textcolor{blue}{[time\_filter\_exact1]} ) > ( SELECT chartevents.valuenum from chartevents WHERE chartevents.icustay\_id IN ( SELECT icustays.icustay\_id from icustays WHERE icustays.hadm\_id IN ( SELECT admissions.hadm\_id from admissions WHERE admissions.subject\_id = \textcolor{teal}{\{patient\_id\}} ) \textcolor{orange}{[time\_filter\_global2]} ) AND chartevents.itemid IN ( SELECT d\_items.itemid from d\_items WHERE d\_items.label = \textcolor{magenta}{\{vital\_name}\} AND d\_items.linksto = `chartevents' ) \textcolor{blue}{[time\_filter\_exact2]} )}
     & \begin{tabular}[t]{@{}p{3in}@{}}
     1. \texttt{SELECT admissions.hadm\_id FROM admissions} \\
    2. \texttt{[PREV\_QUERY] WHERE admissions.subject\_id = \textcolor{teal}{\{patient\_id\}}} \\
    3. \texttt{SELECT icustays.icustay\_id FROM icustays} \\
    4. \texttt{[PREV\_QUERY] WHERE icustays.hadm\_id IN ( [PREV\_RESULT] )} \\
    5. \texttt{[PREV\_QUERY] AND \textcolor{blue}{[time\_filter\_global1]}} \\
    6. \texttt{SELECT d\_items.itemid FROM d\_items} \\
    7. \texttt{[PREV\_QUERY] WHERE d\_items.label = \textcolor{magenta}{\{vital\_name\}} AND d\_items.linksto = 'chartevents'} \\
    8. \texttt{SELECT chartevents.valuenum FROM chartevents} \\
    9. \texttt{[PREV\_QUERY] WHERE chartevents.icustay\_id IN ( [PREV\_RESULT] )} \\
    10. \texttt{[PREV\_QUERY] AND chartevents.itemid IN ( [PREV\_RESULT] )} \\
    11. \texttt{[PREV\_QUERY] \textcolor{blue}{[time\_filter\_exact1]}} \\
    12. \texttt{[PREV\_QUERY] \textcolor{blue}{[time\_filter\_exact2]}} \\
    13. \texttt{SELECT ( [PREV\_RESULT] ) \textcolor{purple}{[comparison]} ( [PREV\_RESULT] )} \end{tabular}
     \\ \hline
    
  \end{tabular}
  }
  \caption{Examples of compositions and components.}
  \label{tab:composition_example}
\end{table*}

%% file: tables/reduced_exec_time.tex
\begin{table*}
  \small 
  \centering 
  \renewcommand{\arraystretch}{1.3}
  \resizebox{1\textwidth}{!}{%
  \begin{tabular}{|C{0.3in}| p{1.2in} | p{3in} | C{0.4in} | p{3in} | C{0.4in}|}
    \hline
    \textbf{Index} & \textbf{Question} & \textbf{Target Representation (SQL$^\dag$)} &  \textbf{Time (SQL$^\dag$)}  & \textbf{Standard SQL} & \textbf{Time (SQL)} \\ \hline
    
    1 & Which icu stay ids are associated with patient 30826 on the current hospital visit?
    & \texttt{SELECT icustays.icustay\_id FROM icustays WHERE icustays.hadm\_id IN ( SELECT admissions.hadm\_id FROM admissions WHERE admissions.subject\_id = 30826 AND admissions.dischtime IS NULL )} & 0.334
    & \texttt{SELECT icustays.icustay\_id FROM icustays WHERE icustays.hadm\_id IN ( SELECT admissions.hadm\_id FROM admissions WHERE admissions.subject\_id = 30826 AND admissions.dischtime IS NULL )} &   0.334 \\ \hline
    
    2& Could you tell me the item id for weight, please?
    & \texttt{SELECT d\_items.itemid FROM d\_items WHERE d\_items.label = `admit wt' AND d\_items.linksto = `chartevents'} & 0.334
    & \texttt{SELECT d\_items.itemid FROM d\_items WHERE d\_items.label = `admit wt' AND d\_items.linksto = `chartevents'} & 0.333 \\ \hline
    
     3 & During result1, what was the last value of result2 that was measured?
     & \texttt{SELECT chartevents.valuenum FROM chartevents WHERE chartevents.icustay\_id IN ( \textcolor{blue}{PREV\_RESULT1} ) AND chartevents.itemid IN ( \textcolor{blue}{PREV\_RESULT2} ) ORDER BY chartevents.charttime DESC LIMIT 1} & 20.352
     & \texttt{SELECT chartevents.valuenum FROM chartevents WHERE chartevents.icustay\_id IN ( SELECT icustays.icustay\_id FROM icustays WHERE icustays.hadm\_id IN ( SELECT admissions.hadm\_id FROM admissions WHERE admissions.subject\_id = 30826 AND admissions.dischtime IS NULL ) ) AND chartevents.itemid IN ( SELECT d\_items.itemid FROM d\_items WHERE d\_items.label = `admit wt' AND d\_items.linksto = `chartevents' ) ORDER BY chartevents.charttime DESC LIMIT 1} & 31.362 \\ \hline

    4 & What about the first measured case?
    & \texttt{SELECT chartevents.valuenum FROM chartevents WHERE chartevents.icustay\_id IN ( \textcolor{blue}{PREV\_RESULT1} ) AND chartevents.itemid IN ( \textcolor{blue}{PREV\_RESULT2} ) ORDER BY chartevents.charttime ASC LIMIT 1} & 19.684
    & \texttt{SELECT chartevents.valuenum FROM chartevents WHERE chartevents.icustay\_id IN ( SELECT icustays.icustay\_id FROM icustays WHERE icustays.hadm\_id IN ( SELECT admissions.hadm\_id FROM admissions WHERE admissions.subject\_id = 30826 AND admissions.dischtime IS NULL ) ) AND chartevents.itemid IN ( SELECT d\_items.itemid FROM d\_items WHERE d\_items.label = `admit wt' AND d\_items.linksto = `chartevents' ) ORDER BY chartevents.charttime ASC LIMIT 1}
    & 30.695 \\ \hline

    5 & What is the variation between result3 and result4?
    & \texttt{SELECT ( \textcolor{blue}{PREV\_RESULT3} ) - ( \textcolor{blue}{PREV\_RESULT4} )} & 0.000
    & \texttt{SELECT ( SELECT chartevents.valuenum FROM chartevents WHERE chartevents.icustay\_id IN ( SELECT icustays.icustay\_id FROM icustays WHERE icustays.hadm\_id IN ( SELECT admissions.hadm\_id FROM admissions WHERE admissions.subject\_id = 30826 AND admissions.dischtime IS NULL ) ) AND chartevents.itemid IN ( SELECT d\_items.itemid FROM d\_items WHERE d\_items.label = `admit wt' AND d\_items.linksto = `chartevents' ) ORDER BY chartevents.charttime DESC LIMIT 1 ) - ( SELECT chartevents.valuenum FROM chartevents WHERE chartevents.icustay\_id IN ( SELECT icustays.icustay\_id FROM icustays WHERE icustays.hadm\_id IN ( SELECT admissions.hadm\_id FROM admissions WHERE admissions.subject\_id = 30826 AND admissions.dischtime IS NULL ) ) AND chartevents.itemid IN ( SELECT d\_items.itemid FROM d\_items WHERE d\_items.label = `admit wt' AND d\_items.linksto = `chartevents' ) ORDER BY chartevents.charttime ASC LIMIT 1 )}
    & 60.722 \\ \hline
    
  \end{tabular}
  }
  \caption{Example of an interaction in EHR-SeqSQL where the target representations (SQL$^\dag$) contain special tokens. We also report average execution times in milliseconds($10^{-3}$), where the queries are executed three times. Their standard SQL versions are also reported for comparison.}
  \label{tab:exec_time_example}
\end{table*}